# Machine Learning-Based Jamun Leaf Disease Detection: A Comprehensive Review

Auvick Chandra Bhowmik

Researcher

Department of Computer Science and Engineering

Daffodil International University, Dhaka, Bangladesh

auvick.bhowmik@yahoo.com

Dr. Md. Taimur Ahad

Associate Professor

Department of Computer Science and Engineering

Daffodil International University, Dhaka, Bangladesh

taimurahad.cse@diu.edu.bd

Yousuf Rayhan Emon

Teaching Assistant

Department of Computer Science and Engineering

Daffodil International University, Dhaka, Bangladesh

yousuf15-3220@diu.edu.bd

*Abstract:* *Jamun leaf diseases pose a significant threat to agricultural productivity, negatively impacting both yield and quality in the jamun industry. The advent of machine learning has opened up new avenues for tackling these diseases effectively. Early detection and diagnosis are essential for successful crop management. While no automated systems have yet been developed specifically for jamun leaf disease detection, various automated systems have been implemented for similar types of disease detection using image processing techniques. This paper presents a comprehensive review of machine learning methodologies employed for diagnosing plant leaf diseases through image classification, which can be adapted for jamun leaf disease detection. It meticulously assesses the strengths and limitations of various Vision Transformer models, including Transfer learning model and vision transformer (TLMViT),*

*SLViT, SE-ViT, IterationViT, Tiny-LeViT, IEM-ViT, GreenViT, and PMViT. Additionally, the paper reviews models such as Dense Convolutional Network (DenseNet), Residual Neural Network (ResNet)-50V2, EfficientNet, Ensemble model, Convolutional Neural Network (CNN), and Locally Reversible Transformer. These machine-learning models have been evaluated on various datasets, demonstrating their real-world applicability. This review not only sheds light on current advancements in the field but also provides valuable insights for future research directions in machine learning-based jamun leaf disease detection and classification.*

**Introduction:**

The Vision Transformer (ViT) stands as a neural network architecture in deep learning designed for image classification purposes (Alzahrani et al., 2023). Functioning as a detection method rooted in pattern recognition and deep learning, Vision Transformer (ViT) possesses the ability to automatically adapt to image features and employ these features for image classification and prediction (Fu et al., 2023). The evolution of transformer architecture, prominently featured in natural language processing (NLP) innovation, has led to the emergence of Vision Transformers (ViT), which, being rooted in Natural Language Processing (NLP), has garnered significant attention in the realm of image classification (Hosseini et al., 2023). In contrast to traditional CNNs that rely on convolution-based architecture, ViT employs a transformer-based architecture, notably effective in tasks related to natural language processing (Alzahrani et al., 2023; Ahmed et al., 2023). ViT aligns with the established data flow pattern of transformers, facilitating its integration with diverse data types (Li et al., 2023). The acyclic network structure of the transformer, coupled with parallel computing through encoder-decoder and self-attention mechanisms, significantly reduces training time and enhances performance in machine translation (Zhan et al., 2023). This marks a significant advancement in visual-based deep learning, where vision transformers have demonstrated substantial promise across tasks extending beyond mere classification (Zhan et al., 2023). Its remarkable ability to simulate long-range dependencies using the self-attention mechanism positions ViT as highly promising in various computer vision applications (Zhang et al., 2021).

In contrast to convolutional neural networks (CNNs), vision transformers (ViTs) are emerging as the predominant technology, gaining popularity for various vision applications (Islam et al., 2022). A pivotal element in leaf disease identification is the Vision Transformer (ViT), an advanced architecture designed for image recognition tasks. Self-attention mechanism band transformers are applied to vision difficulties through the use of vision transformers (ViT), which have been shown to be useful models (Mustofa et al., 2023). ViT incorporates natural language processing and computer vision expertise (Bai et al., 2022). With a focus on long-range correlations, ViTs, a challenging approach in computer vision, have proven effective in addressing various vision problems (Islam et al., 2022). Thanks to its exceptional performance across numerous tasks, Vision Transformer (ViT) is establishing itself as a new frontier in computer vision (Bai et al., 2022). Recent research indicates that, compared to Deep Learning (DL)-based models, transformer-based models are more reliable and yield superior results. Over the past decade, deep learning (DL) has undergone significant transformation due to the rapid advancements in automation and visual analysis technologies (Bai et al., 2022). The utilization of Vision Transformer (ViT) has witnessed a substantial surge in recent times, primarily attributed to its remarkable efficacy in identifying plant leaf diseases. Vision Transformer (ViT) stands out as a key technique to enhance the precision of detecting pests and diseases in jamun leaves.

Jamun fruit, a delectable and nutrient-rich fruit native to Southeast Asia, is also known as jamun, jambolan, or black plum. Jamun fruits contain substantial amounts of iron, calcium, foods rich in calcium, vitamin C, and vitamin A, while being low in fat and calories. Despite its significant health benefits, the productive crop faces challenges due to a prevalent leaf disease affecting jamun production. Factors such as global warming, increased outdoor air pollution, and climate change have contributed to a rise in plant leaf diseases in recent years (Salamai et al., 2023). Plant leaf diseases pose a substantial threat to the integrity of the food supply, and their early detection is challenging, which could mitigate the risk of significant economic harm (De Silva et al., 2023). Implementing appropriate disease control measures is crucial to minimize losses. Diseases like Bacterial Spot, Brown Blight, Coryneum Blight, Powdery Mildew, and Sooty Mold pose severe threats to jamun plants due to their intense symptoms and limited available treatments. Given the significant impact of jamun leaf diseases on both yield and quality, accurate identification is imperative for cultivation and effective treatment. These diseases play a pivotal role in diminishing the productivity of jamun, with various factors such as leaf-eating insects, fluctuating environmental conditions, irregular fertilization, and improper pesticide application contributing to reduced production. Assessing the percentage of damaged leaves in jamun crop fields is vital for determining the appropriate amount of pesticide needed for repairs and predicting changes in crop production. Early detection is crucial for minimizing losses from diseases, and recent technological advancements have been developed to facilitate the identification of plant diseases.

The goal is to establish a dependable model for the detection of jamun leaf diseases in jamun fields, a critical aspect influencing the quality of jamun and the agricultural industry. The identification and segmentation of diseases are pivotal; however, due to the intricate nature of jamun diseases in real-world scenarios, conventional segmentation techniques often struggle to produce rapid and accurate results. Plant disease detection by hand is a more time-consuming process that is only possible in certain locations and is prone to human mistake (Ahad et al., 2023). Deep learning algorithms, particularly Convolutional Neural Networks (CNNs), have shown significant advancements in image classification, object detection, forecasting, segmentation, and various computer vision applications (Sahu et al., 2021). With predictions indicating a global population exceeding 9 billion people by 2050 and a 60% increase in food demand, the detection and appropriate treatment of plant diseases pose challenges for farmers worldwide (Rajput, A. S., 2023). Despite technological advancements in precision agriculture, many existing efforts to classify plant diseases lack effectiveness. Moreover, earlier attempts have struggled to accurately isolate individual leaf segments from entire images, especially in cases where the image background is intricate. In this study, we develop a model to address the challenge of identifying jamun leaf diseases in complex scenes.

## Process of Literature Review:

Studies related to Vision Transformer for detection, feature extraction, feature extraction using traditional machine learning techniques for the classification, autoencoders (AEs), CNN (Convolutional Neural Network), and hybrid models for deep learning techniques included in this study. Most importantly, the research that experimented using a proper research methodology without providing experimental research was included in the literature review.

### Article Identification:

The literature review of this study followed Systematic Reviews and Meta-Analyses (PRISMA) guidelines. A review of around 30 selected papers is presented in this paper. All the articles cover detecting tea leaf disease using Vision Transformer-based deep learning techniques.

In the literature review, the study identified preliminary sources using online databases. Primarily, Google Scholar was selected. The process of article selection is described below:

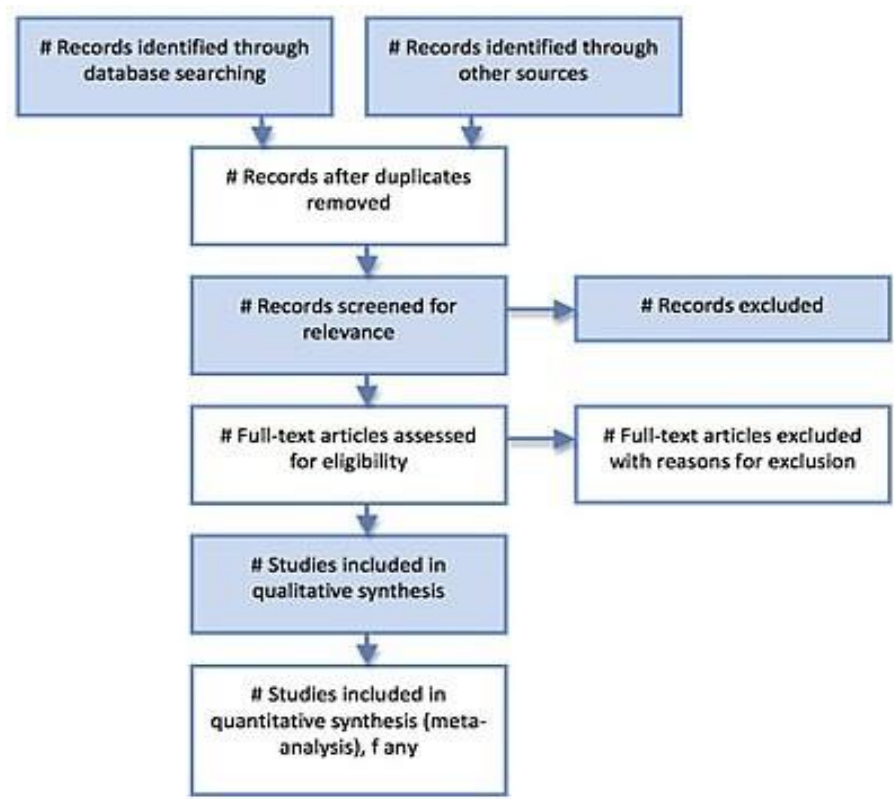

Figure 1: The Process of article selection

**Article Selection:**

Articles were selected for final review using a three-stage screening process based on inclusion and exclusion criteria. After removing duplicate records that were generated from using a database, articles were first screened based on the title alone. The abstract was assessed, and finally, the full articles were checked to confirm eligibility. The chief investigator conducted the entire screening process.

To meet the inclusion criteria, articles had to:

- Be original research articles published in a peer-reviewed journal with full-text access offered within our university;
- Involve the use of plant leaf images;
- Be published in English;
- Be concerned with applying Vision Transformer deep learning techniques for plant leaf disease detection.
- Included articles were limited to those published from 2019 to 2023 to focus on deep learning methodologies. Here, a study was defined as work that employed a Vision Transformer-based deep learning algorithm to detect tea leaf disease and that involved the use of one or more of the following performance metrics: accuracy, the area under the receiver operating characteristics curve, sensitivity, specificity, or F1 score.

The exclusion criteria were:
- Review articles;
- Book or book chapters;
- Conference papers or abstracts;
- Short communications or case reports;
- Unclear descriptions of data;
- No validation was performed.

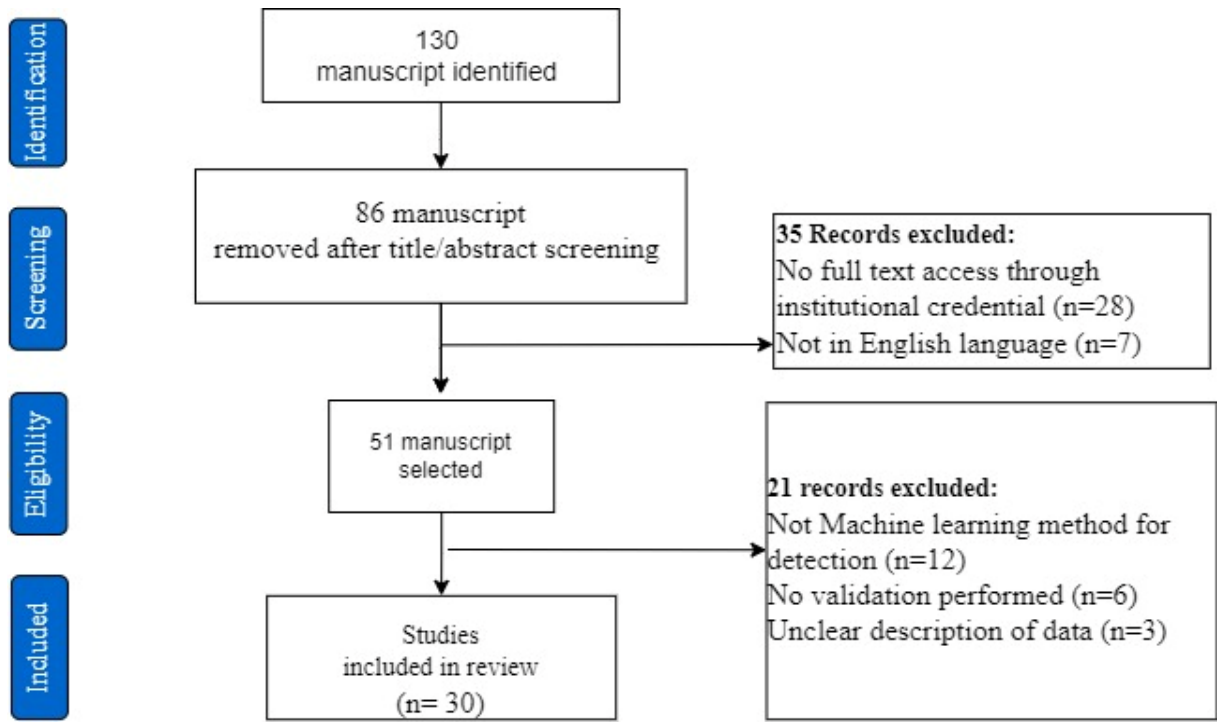

Figure 2: The Preprint Studies

**Literature Review:**

The initial set of researchers employed the conventional Vision Transformer (ViT) model for detecting plant leaf diseases. For instance, Salamai et al. (2023) and De Silva et al. (2023) utilized the traditional ViT model in their studies. Salamai et al. (2023) chose this model due to its ability to creatively learn visual representations of leaf diseases across spatial and channel dimensions. De Silva et al. (2023) opted for this model because of its effectiveness in identifying plant diseases under real environmental conditions. In this line of research, De Silva et al. (2023) achieved a train accuracy of 93.71% and a test accuracy of 90.02%. However, it is noteworthy that both experiments were conducted on publicly available datasets, and conducting experiments on specific datasets related to the actual environment could enhance the precision of these studies.

The second group of researchers employed various hybrid Vision Transformer (ViT) models for the detection of diseases in diverse plant leaves. For instance, Zhou et al. (2023) utilized a pre-trained vision transformer to identify rice leaf diseases, Tabbakh et al. (2023) applied the Transfer Learning Model and Vision Transformer (TLMViT) on a wheat dataset, Li et al. (2023) implemented the Shuffle-convolution-based lightweight Vision Transformer (SLViT) model on sugarcane leaves, and Fu et al. (2023) utilized the Improved Vision Transformer (ViT) for identifying agricultural pests. Additionally, Sun et al. (2023) applied SE-VIT for diagnosing diseases in sugarcane leaves, Zhan et al. (2023) developed the IterationVIT model for diagnosing tea diseases, Thai et al. (2023) used the Tiny-LeViT model for efficient leaf disease detection, and Zhang et al. (2023) applied the IEM-ViT hybrid model for the rapid and accurate recognition of tea diseases. De Silva et al. (2023) employed a ViT + CNN model for plant disease identification, Parez et al. (2023) used GreenViT for identifying plant illnesses, Yu et al. (2023) implemented the Inception convolutional vision transformers model for plant disease identification at inception, Zeng et al. (2023) used the Squeeze-and-Excitation Vision Transformer (SEViT) for identifying large-scale and fine-grained diseases, and Li et al. (2023) implemented the Plant-based MobileViT (PMVT) model for real-time plant disease detection. Several of these studies reported notable accuracies, such as Zhou et al. (2023) achieving 92.00% accuracy, Tabbakh et al. (2023) achieving 98.81% and 99.86% accuracy, Li et al. (2023) achieving a 1.87% accuracy bonus, Sun et al. (2023) achieving 97.26% accuracy, Zhan et al. (2023) achieving 98% accuracy, Thai et al. (2023) achieving 97.25% accuracy, Zhang et al. (2023) achieving 93.78% accuracy, De Silva et al. (2023) achieving 92.83% train accuracy and 88.86% test accuracy, Yu et al. (2023) achieving 99.94%, 99.22%, 86.89%, and 77.54% accuracy, Zeng et al. (2023) achieving 88.34% test accuracy, and Li et al. (2023) achieving 93.6%, 85.4%, and 93.1% accuracy. However, it is worth noting that while Zhou et al. (2023), Zhan et al. (2023), Zhang et al. (2023), and De Silva et al. (2023) conducted their experiments on row datasets, most other researchers implemented their models on publicly available datasets, introducing some limitations to their studies.

The third group of researchers employed various models to conduct comparisons of accuracy and performance on either the same or different datasets. Rethik et al. (2023) utilized ViT1, ViT2, and pre-trained ViT_b16 models to assess accuracy and performance for the classification of plant leaf diseases. The experimentation revealed that the pre-trained ViT_b16 model outperformed the other models. Alzahrani et al. (2023) compared the accuracy of DenseNet169, ResNet50V2, and ViT models for early detection and recognition of tomato leaf diseases, concluding that DenseNet169 demonstrated the highest efficacy. Hossain et al. (2023)

evaluated the accuracy and performance of EANet, MaxViT, CCT, and PVT models for tomato leaf disease detection, with PVT emerging as the superior model. Mustofa et al. (2023) conducted a comparative analysis of ViT, DCNN, CNN, RSNSR-LDD, DDN, and YOLO models for detecting plant leaf diseases, examining their advantages and limitations. Öğrekçi et al. (2023) implemented DenseNet121, ViT, and ViT + CNN models to compare accuracy and performance in identifying sugarcane leaf diseases, with ViT proving to be the most effective model. In these studies, some researchers achieved notable accuracy for robust model comparisons. For instance, Rethik et al. (2023) attained accuracy values of 85.87%, 89.16%, and 94.16% for ViT1, ViT2, and pre-trained ViT_b16 models, respectively. Alzahrani et al. (2023) reported 99.88% train and 99.00% test accuracy for the DenseNet121 model, and 95.60% train and 98.00% test accuracy for the ResNet50V2 and ViT models. Hossain et al. (2023) found accuracy rates of 89%, 97%, 91%, and 93% for EANet, MaxViT, CCT, and PVT models, respectively. Öğrekçi et al. (2023) achieved accuracy values of 92.87%, 93.34%, and 87.37% for DenseNet121, ViT, and ViT + CNN models, respectively. It's worth noting that all researchers implemented their models on publicly available datasets, which, among other factors, may limit the extent to which their studies contribute to knowledge enrichment compared to the use of more specialized datasets.

The fourth group of researchers employed multiple models to construct ensemble models, aiming for improved accuracy and performance. Kumar et al. (2023) utilized EfficientNet, SEResNeXt, ViT, DeIT, and MobileNetV3 models to develop an ensemble model for the detection of cassava leaf diseases. Ganguly et al. (2023) introduced an ensemble model that incorporates CNN, ResNeXt, and InceptionV3 for the detection of plant leaf diseases. Chang et al. (2024) constructed an ensemble model comprising ViT, PVT, and Swin to enhance the quality of identification in plant leaf disease detection. Among these studies, only Kumar et al. (2023) achieved an accuracy of 90.75% on their ensemble model. However, it is notable that all other researchers conducted their experiments on publicly available datasets, introducing some limitations to their studies. Implementing experiments on specific datasets could potentially enhance the precision and richness of these studies.

The final group of researchers employed diverse models to enhance accuracy and performance in disease detection. Diana et al. (2023) utilized convolutional capsule networks for identifying grape leaf diseases. Kumar et al. (2023) introduced paddy leaf disease detection using a multi-scale feature fusion-based RDTNet. Hu et al. (2023) proposed an Adaptive Fourier Neural Operators (AFNO)-based Transformer architecture named FOTCA, focusing on extracting global features in advance. Arshad et al. (2023) presented a novel hybrid deep learning model, PLDPNet, designed for automatic forecasting of potato leaf diseases. Zhang et al. (2023) introduced a unique segmentation model for grape leaf diseases in natural scene photos, termed Locally Reversible Transformer (LRT). Thai et al. (2023) developed a transformer-based leaf disease detection model known as FormerLeaf. Devi et al. (2023) created an EffectiveNetV2 model for pest identification and plant disease categorization. Diana et al. (2023) achieved 99.12% accuracy in their study. Kumar et al. (2023) attained 99.55%, 99.54%, and 99.53% accuracy, f1-score, and precision, respectively. Hu et al. (2023) reported a 99.8% accuracy for the FOTCA model. Arshad et al. (2023) achieved an overall accuracy of 98.66% with their proposed model. Thai et al. (2023) claimed a 3% accuracy improvement in disease detection with their model FormerLeaf. Devi et al. (2023) reached 99.5%, 97.5%, and 80.1% accuracy in their study. Given the use of publicly available datasets by all researchers, there may be a potential deficiency in knowledge enrichment in their studies at some level.

**Research Matrix:**

| Author | Model | Accuracy | Contribution |
| --- | --- | --- | --- |
| Salamai et al. (2023) | ViTs |  | The visual modulation network can innovatively learn the visual representations of leaf. |
| De Silva et al. (2023) | ViT | Train 93.71 % and Test 90.02 % | This work highlights the potential of utilizing advanced imaging techniques for accurate and reliable plant disease identification. |

| Zhou et al. (2023) | Pre-trained vision transformer | 92.00% | Proposed a residual-distilled transformer architecture. |
| Tabbakh et al. (2023) | Transfer learning model and vision transformer (TLMViT) | 98.81% & 99.86% | Proposed a hybrid model for plant disease classification. |
| Li et al. (2023) | Shuffle-convolution-based lightweight Vision transformer (SLViT) | Accuracy bonus of 1.87 % over MobileNetV3_small | Present a hybrid model that initially trained on the publicly available dataset |
| Fu et al. (2023) | Improved Vision Transformer (ViT) | | This model has high accuracy of image processing & recognition technology. |
| Sun et al. (2023) | SE-VIT | 97.26% | SE-VIT can provide smart management of sugarcane plantations and addressing plant diseases with limited datasets. |
| Zhan et al. (2023) | IterationVIT | 98% | IterationVIT model can accurately capture the location of diseases. |
| Thai et al. (2023) | Tiny-LeViT | 97.25% | Proposed Tiny-LeViT model based on the transformer architecture for efficient leaf disease classification. |
| Zhang et al. (2023) | IEM-ViT | 93.78% | In comparison ResNet18, VGG16, and VGG19, the recognition |

| | | | accuracy was improved by nearly 20%. |
|---|---|---|---|
| De Silva et al. (2023) | ViT + CNN | Train 92.83% & Test 88.86% | Hybrid ViT combine the strengths of both CNN and ViT. |
| Parez et al. (2023) | GreenViT | | Proposed technique outperforms state-of-the-art (SOTA) CNN models for detecting plant diseases. |
| Yu et al. (2023) | Inception convolutional vision transformers | 99.94, 99.22%, 86.89% and 77.54% | Inception architecture & cross channel feature learning can improve the information richness that beneficial to fine-grained feature learning. |
| Zeng et al. (2023) | Squeeze-and-Excitation Vision Transformer (SEViT) | Test 88.34% | Compared with the baseline model, the classification accuracy of SEViT is improved by 5.15%. |
| Li et al. (2023) | Plant-based MobileViT (PMVT) | 93.6%, 85.4% & 93.1% | Developed a plant disease diagnosis app using PMVT model to identify plant disease in different scenarios. |
| Rethik et al. (2023) | ViT1, ViT2 & pre-trained ViT_b16 | 85.87%, 89.16% and 94.16%. | Researchers have replaced CNN with a new method called Vision Transformer for classifying plant leaf diseases. |

| Alzahrani et al. (2023) | DenseNet169, ResNet50V2 & ViT | (Train 99.88% and Test 99.00%) & (Train 95.60% and Test 98.00%) | Compared the performance of three different deep learning models. |
|---|---|---|---|
| Hossain et al. (2023) | EANet, MaxViT, CCT & PVT | 89%, 97%, 91% & 93% | Analyze the effect of four different transformer-based models. |
| Mustofa et al. (2023) | ViT, DCNN, CNN, RSNSR-LDD, DDN, YOLO | | The advantages and limitations of different deep learning models are described. |
| Öğrekçi et al. (2023) | DenseNet121, ViT, ViT + CNN | 92.87%, 93.34% & 87.37% | Compare among three model for best output. |
| Kumar et al. (2023) | EfficientNet, SEResNeXt, ViT, DeIT & MobileNetV3 | 90.75% | Proposes an ensemble algorithm a few state-of-the-art models in classification and object detection. |
| Ganguly et al. (2023) | Ensemble model (CNN, ResNeXt, and InceptionV3) | | The proposed approach contributes to the advancement of automated plant disease diagnosis systems. |
| Chang et al. (2024) | ViT, PVT, and Swin | | These models enhance the fusion of multiscale features and edge information. |
| Diana et al. (2023) | Convolutional capsule networks | 99.12% | The performance is compared with state-of-the-art deep learning methods and produces reliable results. |

| Kumar et al. (2023) | RDTNet | 99.55%, 99.54% & 99.53% | RDTNet has two modules that reduce computation cost & improve accuracy. |
| --- | --- | --- | --- |
| Hu et al. (2023) | FOTCA | 99.8% | Proposed a hybrid transformer-CNN architecture using AFNO. |
| Arshad et al. (2023) | PLDPNet | 98.66% | Developed a hybrid deep-learning framework & employ ensemble approach to get powerful features. |
| Zhang et al. (2023) | Locally Reversible Transformer (LRT) | | Segmentation performance of LRT outperforms state-of-the-art models with comparable GFLOPs and parameters. |
| Thai et al. (2023) | FormerLeaf | Enhance 3% accuracy | Using two optimization methods LeIAP & SPMM to enhance the model performance. |
| Devi et al. (2023) | EfficientNetV2 | 99.5%, 97.5%, 80.1% | EfficientNetV2 has a faster rate irrespective of the data size and class distribution as compared to InceptionV3 and ViT models. |

**Inference from current research studies:**

As indicated in the research matrix, there has been a recent surge in research on plant disease detection, driven by the integration of state-of-the-art machine learning and transformer-based models. In the realm of detecting diseases in plant leaves, various scholars have undertaken diverse research endeavors, employing a broad spectrum of methodologies and models. The initial cohort of researchers, exemplified by Salamai et al. (2023) and De Silva et al. (2023), has underscored the potential of traditional Vision Transformers (ViTs) due to their remarkable precision and adaptability to multispectral data. While De Silva et al. (2023) achieved a commendable accuracy of 93.71% in training and 90.02% in testing, a predominant concern arises regarding their limited focus on plant leaf datasets, raising questions about the broader applicability of ViTs in this specific domain. Conversely, other researchers such as Zhou et al. (2023), Tabbakh et al. (2023), Fu et al. (2023), and others consistently favored advanced ViTs in the context of plant disease detection, attaining noteworthy validation accuracy. However, the absence of experiments conducted on jamun leaf datasets within this category introduces uncertainties regarding the suitability of these models for jamun leaf disease detection.

The integration of Vision Transformers with other models is employed to enhance the efficiency and precision of disease detection. Notable researchers in this domain, including Li et al. (2023) and Yu et al. (2023), have demonstrated promising outcomes within the context of standard plant disease datasets. However, the absence of a primary dataset and a specific focus on jamun leaf diseases in these studies underscores the need for additional investigation. Concurrently, researchers have made substantial contributions to the field by developing deep Convolutional Neural Network (CNN) models specifically designed for plant leaf diseases. As exemplified by authors such as Sun et al. (2023), Zhan et al. (2023), Thai et al. (2023), and others, these models exhibit impressive accuracy in identifying various plant leaf diseases. Nevertheless, their applicability in a diverse range of real-world farming scenarios remains an aspect that necessitates further exploration.

Moreover, the examination of existing literature highlights the potential of Ensemble-based models in object detection for identifying diverse plant leaf diseases. This is evident in research conducted by Ganguly et al. (2023) and other studies, which have showcased outstanding accuracy and precision in detecting these diseases. However, a careful examination is required to assess the adaptability of these models to real-world datasets. Furthermore, a cohort of researchers has introduced innovative techniques and technologies, including MobileNet, Swin Transformer, and EfficientNetV2, demonstrating commendable accuracy across various agricultural applications. Nonetheless, the absence of experiments utilizing primary datasets underscores the necessity for more comprehensive investigations into the practicality and effectiveness of these models in detecting different plant leaf diseases. In summary, these insights underscore the dynamic nature of research in plant leaf disease detection and emphasize the crucial role of future studies in addressing the unique challenges and datasets relevant to this pivotal field.

**Limitations:**

This study aims to provide an in-depth evaluation of the current status of detecting diseases in jamun leaves by employing advanced deep learning techniques. Furthermore, the research seeks to explore a less-explored avenue within this domain, specifically focusing on the use of Vision Transformers to mitigate losses in jamun yield and quality. This research direction aligns closely with the broader goal of leveraging technological advancements to enhance disease management strategies and elevate jamun production.

It's important to highlight that a substantial portion of the current literature focuses on employing publicly accessible datasets for training and assessing models dedicated to disease detection in various plant leaves. Many studies primarily rely on a single dataset, raising uncertainties about the model's generalizability to other datasets. Additionally, it is essential to recognize that the success of deep learning approaches in disease classification is contingent on access to large, well-annotated image datasets. Obtaining a diverse and comprehensive dataset with accurately labeled instances of different diseases presents a significant challenge.

Generating datasets of this nature is a complicated and time-intensive undertaking. The various forms of jamun leaf diseases, coupled with factors like lighting conditions and growth stages, introduce variability in their appearance, posing a challenge in constructing a dependable dataset. Training deep learning models effectively necessitates diverse and high-quality training data, a task made difficult by these complexities. Specialists in the study of jamun leaf diseases encounter the need to collect numerous images with precise labels to identify different diseases within each image. This demands expertise and a substantial time commitment. Consequently, many researchers opt to utilize existing datasets that are accessible to the broader community.

**Direction for Future Research:**

Future research endeavors aiming to enhance the accuracy and efficiency of detecting jamun leaf diseases using machine learning encompass reinforcement learning, hybrid machine learning, and case-based reasoning. Reinforcement learning holds potential for developing a Vision Transformer (ViT) model capable of autonomously identifying diseased jamun leaves, eliminating the need for human intervention. Hybrid machine learning involves integrating multiple models with a traditional machine learning algorithm to enhance the precision of jamun leaf disease detection. Case-based reasoning offers the opportunity to build a machine learning model that learns from the errors of previous models, progressively improving its performance.

Alongside these innovative research directions, several challenges must be addressed to enhance the accuracy and efficiency of jamun leaf disease detection using ViT. These challenges encompass issues such as data scarcity, environmental variability, and real-time detection. Despite these hurdles, ViT remains a promising technology for jamun leaf disease detection. By persistently exploring novel techniques and tackling existing challenges, it is conceivable to develop ViT models that can adeptly and efficiently identify jamun leaf diseases.

**Conclusion:**

This study provides a comprehensive overview of the detection of jamun leaf diseases, delving into diverse approaches and methodologies for the early diagnosis and management of plant diseases. Vision Transformers (ViT) have emerged as a promising technology for precisely classifying and identifying jamun leaf diseases, alongside other models such as deep convolutional neural networks (CNN), Ensemble-based techniques, and hybrid models. Researchers have scrutinized the merits and limitations of these models, conducting thorough assessments using metrics like accuracy, precision, and recall. While ViT-enabled CNN models and other innovative techniques exhibit promise, it is crucial to consider their practical utility in real-world scenarios, particularly concerning jamun leaf datasets. Furthermore, the efficiency of ensembled models that integrate vision transformers with other approaches has been demonstrated, and advanced CNN models and Ensemble-based models have proven effective in recognizing various jamun leaf diseases. These collective findings provide valuable insights into the progression of jamun leaf disease detection, with a dedicated focus on enhancing agricultural practices and addressing the specific challenges encountered by farmers in practical settings.